\documentclass[runningheads]{llncs}
\usepackage[T1]{fontenc}
\usepackage{graphicx}
\usepackage{CJK}

\usepackage{float}
\usepackage{url}
\usepackage{enumitem}
\usepackage{todonotes}
\usepackage{booktabs}
\usepackage{dashrule}
\usepackage{mdframed}
\usepackage{listings}
\usepackage[normalem]{ulem}
\usepackage{subcaption}
\usepackage{tikz}
\usetikzlibrary{positioning}
\usepackage{xcolor}
\usepackage{setspace}
\usepackage{amsmath}
\newenvironment{claimreasoningexample}{%
\vspace{0.5mm}
\begin{mdframed}[backgroundcolor=yellow!10]
\small 
}{%
\end{mdframed}
\vspace{0.5mm}
}

\renewcommand{\claim}[1]{\par\noindent\textit{\textbf{Claim:}} #1}
\newcommand{\reasoning}[1]{\par\noindent\textit{\textbf{Reasoning:}} #1}

\begin{document}
\begin{CJK*}{UTF8}{gbsn}
\title{What Do Claim Verification Datasets Actually Test? \\ A Reasoning Trace Analysis}
\titlerunning{Claim Verification Dataset Analysis}

\author{Delip Rao\orcidID{0000-0002-4534-9906} \and
Chris Callison-Burch\orcidID{0000-0001-8196-1943}}

\authorrunning{D. Rao and C. Callison-Burch}

%
\institute{University of Pennsylvania, Philadelphia, PA, USA \\
\email{\{delip, ccb\}@seas.upenn.edu}}

\maketitle              

\begin{abstract}
Despite rapid progress in claim verification, we lack a systematic understanding of what reasoning these benchmarks actually exercise. We generate structured reasoning traces for 24K claim-verification examples across 9 datasets using GPT-4o-mini and find that direct evidence extraction dominates, while multi-sentence synthesis and numerical reasoning are severely under-represented. A dataset-level breakdown reveals stark biases: some datasets almost exclusively test lexical matching, while others require information synthesis in roughly half of cases. Using a compact 1B-parameter reasoning verifier, we further characterize five error types and show that error profiles vary dramatically by domain -- general-domain verification is dominated by lexical overlap bias, scientific verification by overcautiousness, and mathematical verification by arithmetic reasoning failures. Our findings suggest that high benchmark scores primarily reflect retrieval-plus-entailment ability. We outline recommendations for building more challenging evaluation suites that better test the reasoning capabilities verification systems need.

\end{abstract}

\begin{figure}[h!]
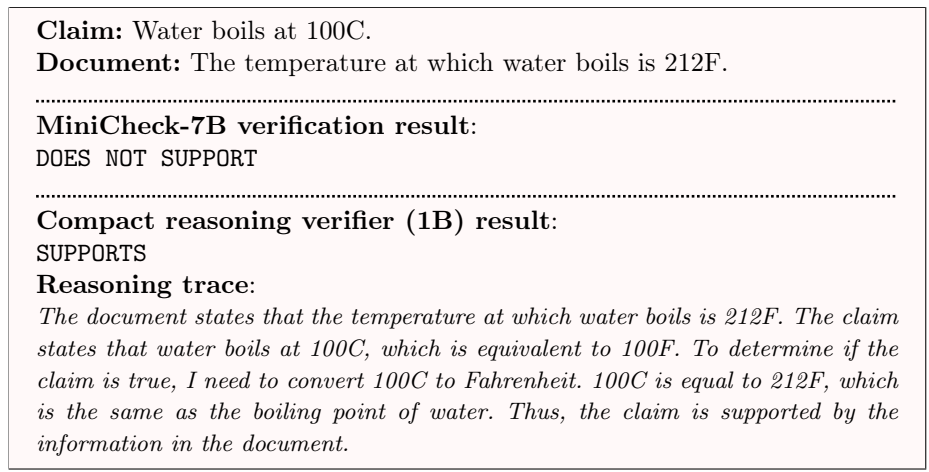

    \centering
    \begin{mdframed}[backgroundcolor=pink!10]
    \noindent \textbf{Claim:} Water boils at 100C.\\
    \noindent \textbf{Document:} The temperature at which water boils is 212F.\\
    \noindent\hdashrule{\textwidth}{1pt}{1pt}
    \noindent \textbf{MiniCheck-7B verification result}:\\ \texttt{DOES NOT SUPPORT}\\
    \noindent\hdashrule{\textwidth}{1pt}{1pt}\\
    \noindent \textbf{Compact reasoning verifier (1B) result}:\\ \texttt{SUPPORTS}\\
    \noindent \textbf{Reasoning trace}:\\ \textit{\footnotesize The document states that the temperature at which water boils is 212F. The claim states that water boils at 100C, which is equivalent to 100F. To determine if the claim is true, I need to convert 100C to Fahrenheit. 100C is equal to 212F, which is the same as the boiling point of water. Thus, the claim is supported by the information in the document.}
    \end{mdframed}
    \vspace{-0.9em}
    \caption{Surface matching fails on claims requiring reasoning. A state-of-the-art 7B verifier~\cite{tang2024minicheck} rejects the claim because $100^{\circ}C$ does not appear in the document, even though $100^{\circ}C = 212^{\circ}F$. A compact reasoning verifier that generates an explicit reasoning trace before deciding correctly handles the unit conversion. This example motivates our analysis: if benchmarks primarily reward surface matching, high scores may not reflect genuine verification ability.}
    \vspace{-1.2em}
    \label{fig:motivating-example}
\end{figure}

\section{Introduction}
\label{sec:introduction}

Claim verification benchmarks such as LLMAggreFact~\cite{tang2024minicheck}, SciFact~\cite{wadden-etal-2020-fact}, and FEVER~\cite{thorne2018fever} are widely used to evaluate factual grounding models. Recent systems report near-human balanced accuracies on these benchmarks, yet practical deployment reveals persistent failures on claims requiring multi-step reasoning, numerical computation, or evidence synthesis across sentences (see Figure~\ref{fig:motivating-example} for a simple illustration). This gap raises a question: what kinds of reasoning do current benchmarks actually exercise?

To answer this question, we generate structured reasoning traces for 30.4K examples from the LLMAggreFact benchmark -- which aggregates nine widely-used claim-verification datasets -- using GPT-4o-mini in a zero-shot setting. After filtering for label agreement with ground truth, we retain 24.1K traces and systematically analyze the reasoning strategies they reveal. We complement this with an error analysis using a compact 1B-parameter reasoning verifier that we train on these traces, which allows us to characterize failure modes across general, scientific, and mathematical domains.

Our contributions are:
\begin{itemize}[noitemsep,topsep=0pt]
\item A taxonomy of six reasoning patterns observed in claim verification, based on manual review of 1{,}000 stratified samples.
\item Quantitative evidence that direct evidence extraction (surface matching) accounts for the majority of verification reasoning, while multi-sentence synthesis and step-by-step verification are rare.
\item A dataset-level analysis revealing significant variation in reasoning demands across the nine constituent datasets in LLMAggreFact.
\item An error taxonomy from a compact verifier identifying five domain-dependent error types, showing that error profiles differ dramatically between general, scientific, and mathematical claims.
\item Concrete recommendations for constructing more challenging claim verification benchmarks.
\end{itemize}

\section{Related Work}
\label{sec:related_work}

\paragraph{Dataset bias and annotation artifacts.}
Studies in NLI have shown that benchmarks can contain systematic biases that allow models to achieve high accuracy without genuine understanding~\cite{pmlr-v139-zhao21c}. Similar concerns apply to claim verification, where surface-level cues may substitute for deeper reasoning.

\paragraph{Claim verification benchmarks.}
LLMAggreFact~\cite{tang2024minicheck} aggregates nine sources spanning news, science, and dialogue domains~\cite{tang2022understanding,nallapati2016abstractive,narayan2018don,zhu2021mediasum,hu2023meetingbank,liu2023evaluating,malaviya2023expertqa,wang2023factcheck,kamoi2023wice}. SciFact~\cite{wadden-etal-2020-fact} and FEVER~\cite{thorne2018fever} target scientific and Wikipedia claims. While these benchmarks have driven progress, the reasoning complexity they demand has not been systematically characterized.

\paragraph{Reasoning in NLP evaluation.}
Chain-of-thought prompting~\cite{wei2022chain} has enabled analysis of model reasoning processes. Recent work on verifiable CoT~\cite{jacovi2024chain} and reasoning-augmented verification~\cite{yao2023react} provides tools for inspecting what reasoning models actually perform. Large reasoning models such as DeepSeek R1~\cite{deepseek-r1-2025} generate verbose traces, while distilled approaches can produce more concise rationales suitable for systematic analysis.

\paragraph{Lightweight verifiers.}
Compact verification models such as MiniCheck~\cite{tang2024minicheck} and AlignScore~\cite{zha2023alignscore} have shown that specialized training can match larger models. We use a compact reasoning verifier as an analysis tool to characterize error patterns across domains.

\section{Reasoning Trace Generation}
\label{sec:trace-generation}

To systematically analyze the reasoning demands of claim verification benchmarks, we generated structured reasoning traces for all 30.4K examples in the LLMAggreFact development set. Using zero-shot prompting, GPT-4o-mini generated a step-by-step reasoning process and a YES/NO verification label for each (document, claim) pair. To ensure trace quality, we filtered instances where the generated label mismatched the original LLMAggreFact label, reducing the dataset from 30.4K to 24.1K examples.\footnote{Approximately 21\% of LLMAggreFact dev set labels differed from GPT-4o-mini's predictions; analyzing this discrepancy is beyond this paper's scope.} This filtered set contains traces across all nine constituent datasets in LLMAggreFact. We randomly sampled 100 traces across all 9 datasets and manually verified their accuracy. We opted against DeepSeek R1~\cite{deepseek-r1-2025} traces due to their verbosity and token inefficiency, which would complicate systematic analysis.

\section{Reasoning Pattern Taxonomy}
\label{sec:reasoning_patterns_taxonomy}

\subsection{Methodology}
We sampled 1{,}000 instances with stratified sampling over the dataset source, then manually reviewed the generated rationales for each (document, claim) pair. For every instance, annotators identified the primary reasoning strategy used to justify the predicted label. Through an iterative pass, we consolidated categories, recorded recurring patterns, and selected representative examples (Section~\ref{sec:examples}).

\subsection{Reasoning Patterns}
\label{subsec:reasoning_patterns}
We observe six recurring strategies: (A) \emph{Direct evidence extraction \& matching} (quoting or paraphrasing spans that explicitly support/contradict the claim); (B) \emph{Handling nuance \& implication} (contextual or partial support without verbatim statements); (C) \emph{Absence of evidence identification} (stating that the document lacks the requisite information); (D) \emph{Synthesis of multiple information points} (integrating evidence across sentences/sections); (E) \emph{Addressing scope/specificity mismatches} (claim broader/narrower or adding elements absent from the source); and (F) \emph{Step-by-step verification} (checking procedural or sequential claims).

\subsection{Overall Distribution}

\begin{figure}[h]
    \centering
    \includegraphics[width=1\linewidth]{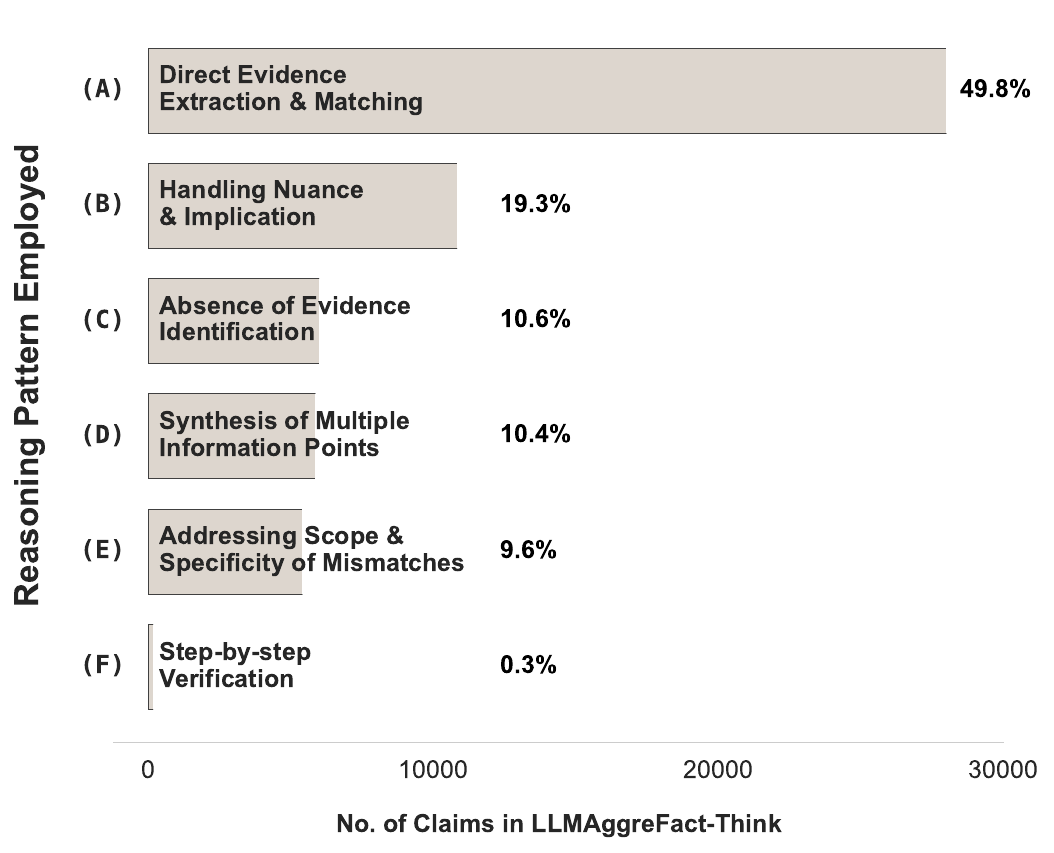}
    \vspace{-1.0em}
    \caption{Distribution of reasoning patterns across 24.1K claim verification traces. Direct evidence extraction (A) dominates the verification strategies (27,988 instances), followed by other reasoning strategies. See Section~\ref{subsec:reasoning_patterns} for pattern definitions.}
    \vspace{-1.0em}
    \label{fig:reasoning_methods_distribution}
\end{figure}

Overall distribution (Figure~\ref{fig:reasoning_methods_distribution}) shows a strong dominance of direct extraction (A), with substantially smaller shares for nuance (B), absence identification (C), and multi-sentence synthesis (D); step-by-step verification (F) is rare. This indicates that the majority of verification reasoning in LLMAggreFact reduces to locating and matching relevant text spans.

\subsection{Dataset-Level Analysis}

\begin{figure}[h]
    \centering
    \includegraphics[width=1\linewidth]{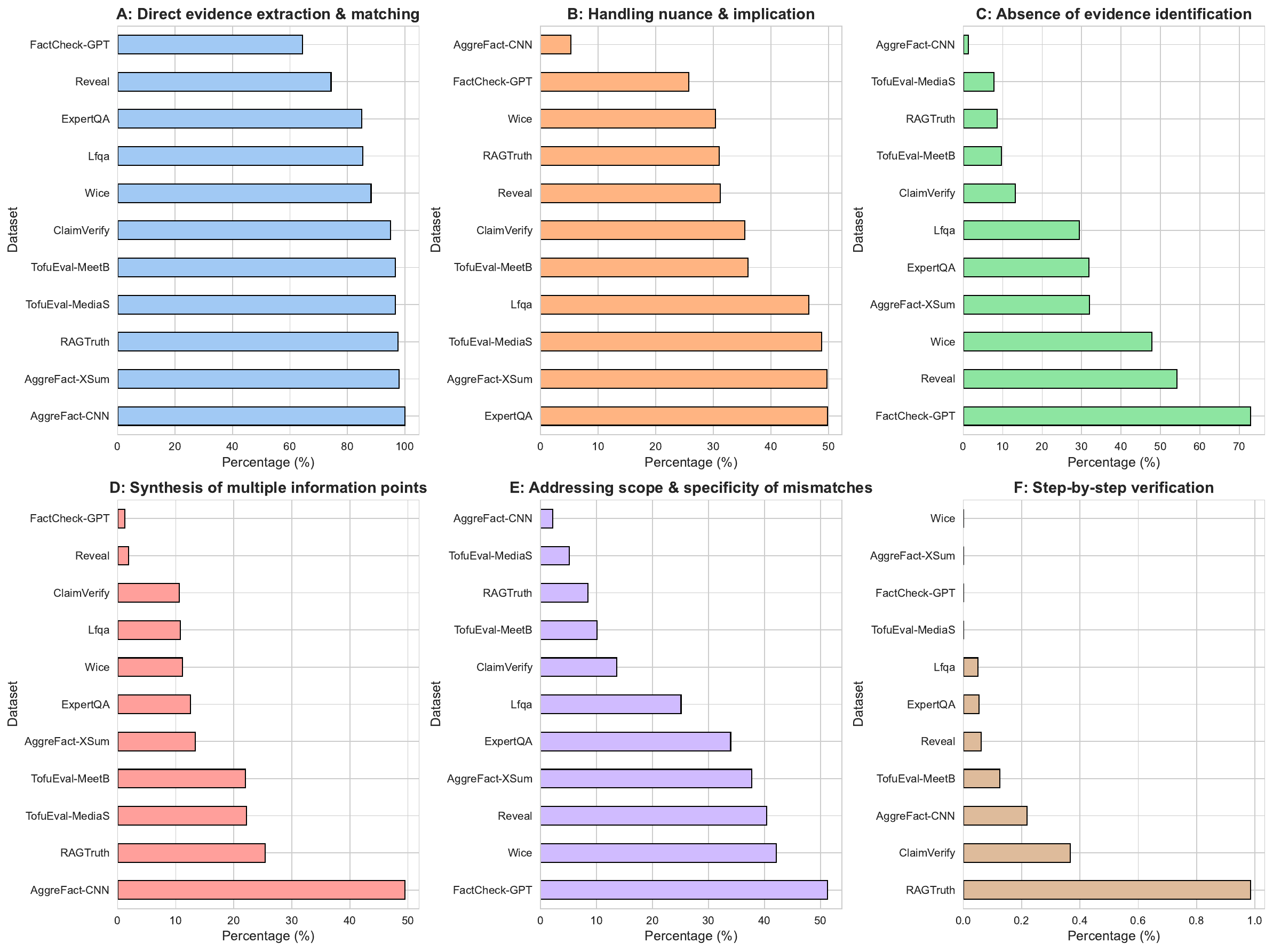}
    \vspace{-1.0em}
    \caption{Distribution of reasoning strategies across the nine source datasets in LLMAggreFact. Each subplot (A--F) corresponds to a reasoning pattern: (A) Direct evidence extraction \& matching, (B) Handling nuance \& implication, (C) Absence of evidence identification, (D) Synthesis of multiple information points, (E) Addressing scope \& specificity mismatches, and (F) Step-by-step verification. Different datasets elicit markedly different reasoning patterns.}
    \vspace{-1.0em}
    \label{fig:dataset_method_distributions}
\end{figure}

The dataset-level breakdown (Figure~\ref{fig:dataset_method_distributions}) reveals clear biases: e.g., AggreFact-CNN requires information synthesis in roughly half of cases, whereas even the dataset with the highest rate of multi-step verification (RAGTruth) exhibits it in only $\sim$0.9\% of instances. We also observe complementarity between patterns: sources with very high direct extraction (A) tend to have low synthesis (D). Scope/nuance phenomena (E/B) are more prominent in FactCheck-GPT and Wice, while absence identification (C) varies drastically by formulation (from $<1$\% in AggreFact-CNN to $>70$\% in FactCheck-GPT).

\section{Representative Examples}
\label{sec:examples}

We present one example for each identified reasoning pattern, drawn from the generated traces.

\paragraph{Direct Evidence Extraction \& Matching:}
\begin{claimreasoningexample}
    \claim{"Proper clean-up prior to insertion of the vaginoscope is essential."}
    \reasoning{"The document contains a sentence that explicitly states 'Proper clean-up prior to insertion of the vaginoscope is essential.' This directly matches the claim made..."}
\end{claimreasoningexample}

\paragraph{Absence of Evidence Identification:}
\begin{claimreasoningexample}
    \claim{"A star's temperature is determined by the amount of mass and energy it has."}
    \reasoning{"The document discusses determining the effective temperature of a star from its spectrum but does not mention the relationship between a star's mass, energy, and temperature... Therefore, the claim is not backed by the facts in the document..."}
\end{claimreasoningexample}

\paragraph{Synthesis of Multiple Information Points:}
\begin{claimreasoningexample}
    \claim{"However, according to customer reviews, the quality of the food served at the restaurant is subpar."}
    \reasoning{"The document contains multiple customer reviews expressing dissatisfaction... One review explicitly states that the food was 'horrible,'... Another review criticizes the broth as 'least flavorful'... Together, these reviews confirm that the quality of food... is indeed regarded as subpar by customers."}
\end{claimreasoningexample}

\paragraph{Addressing Scope and Specificity Mismatches:}
\begin{claimreasoningexample}
    \claim{"Thermodynamics provides the fundamental laws of energy conservation, mass conservation, and chemical equilibrium that govern the behavior of chemical systems."}
    \reasoning{"The document discusses the first law of thermodynamics, which addresses energy conservation, aligning with part of the claim. However, it does not support the components of mass conservation and chemical equilibrium... Therefore, the claim goes beyond what is explicitly stated... and is only partially supported."}
\end{claimreasoningexample}

\paragraph{Handling Nuance and Implication:}
\begin{claimreasoningexample}
    \claim{"The restaurant's use of fresh and high-quality ingredients, such as shrimp and pork belly, has also been commended by reviewers."}
    \reasoning{"...One review even states that 'the shrimp you could really sink your teeth into,' suggesting that the shrimp is fresh and of high quality. The combination of positive comments about these specific ingredients supports the claim."}
\end{claimreasoningexample}

\paragraph{Step-by-Step Verification:}
\begin{claimreasoningexample}
    \claim{"Return beef and any accumulated juices to the pan, along with tomato sauce, liver spread, and bay leaves."}
    \reasoning{"...passage 3 explicitly states to 'Return beef and accumulated juices to pan' and also adds 'Add water, tomato sauce, liver spread and bay leaf.' Therefore, all elements of the claim are present in the passage..."}
\end{claimreasoningexample}

\section{Error Analysis from a Compact Verifier}
\label{sec:error-analysis}

\subsection{Approach}
To characterize where current verification systems fail, we trained ClaimTrace, a compact reasoning verifier, on the generated traces. ClaimTrace is a 4-bit quantized Gemma3 1B model~\cite{gemmateam2025gemma3technicalreport} fine-tuned with LoRA~\cite{hu2021lora} to produce structured reasoning followed by a verification decision. We evaluate ClaimTrace on held-out data from three domains: general claims (LLMAggreFact), scientific claims (SciFact~\cite{wadden-etal-2020-fact}), and mathematical claims (a math-reasoning benchmark derived from GSM8K~\cite{cobbe2021}), and systematically categorize its errors.

\subsection{Error Taxonomy}
Through systematic analysis of ClaimTrace predictions across all three domains, we identified five primary error categories (six for mathematical claims):
\textbf{Lexical Overlap Bias}: incorrectly predicting support based on surface-level lexical similarity without proper semantic entailment.
\textbf{Insufficient Aggregation}: failure to synthesize information distributed across multiple sentences.
\textbf{Negation/Temporal Confusion}: mishandling negations or temporal relationships.
\textbf{Overcautiousness}: requiring complete explicit evidence for all components of a claim, defaulting to rejection even when most sub-claims are supported.
\textbf{Hallucinated Justification}: generating confident reasoning unsupported by the document.
For mathematical claims, we add \textbf{Arithmetic Reasoning}: failure to execute correct computation despite understanding the problem structure.

\subsection{Error Distribution across Domains}

\begin{figure*}[h!]
    \centering
    \begin{subfigure}[b]{0.32\textwidth}
        \includegraphics[width=\textwidth]{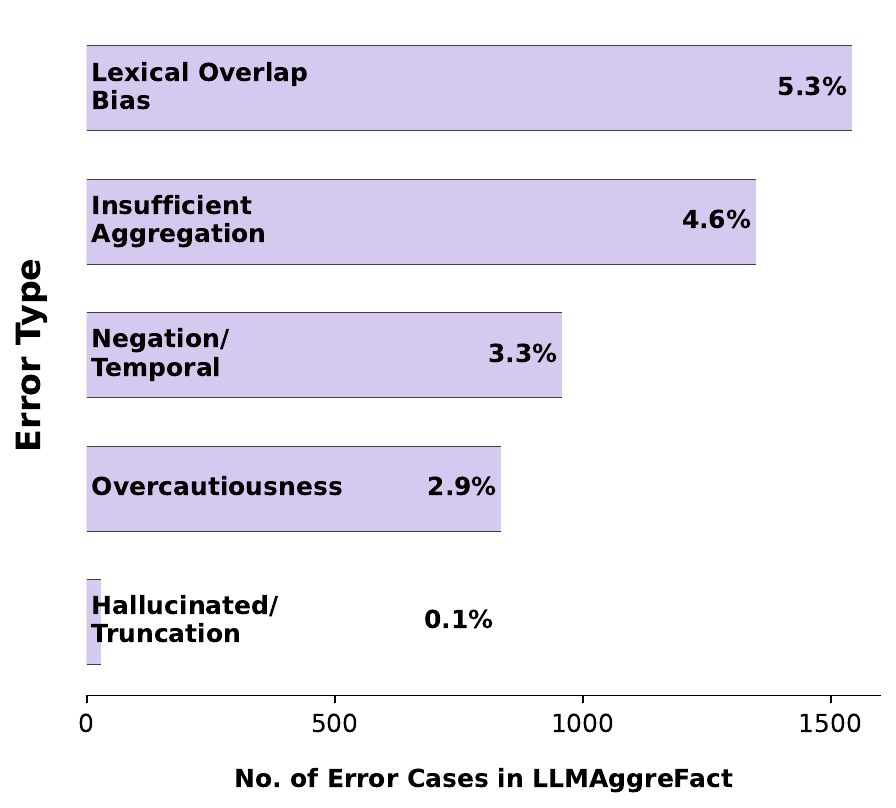}
        \caption{LLMAggreFact}
        \label{fig:error_llmaggrefact}
    \end{subfigure}
    \hfill
    \begin{subfigure}[b]{0.32\textwidth}
        \includegraphics[width=\textwidth]{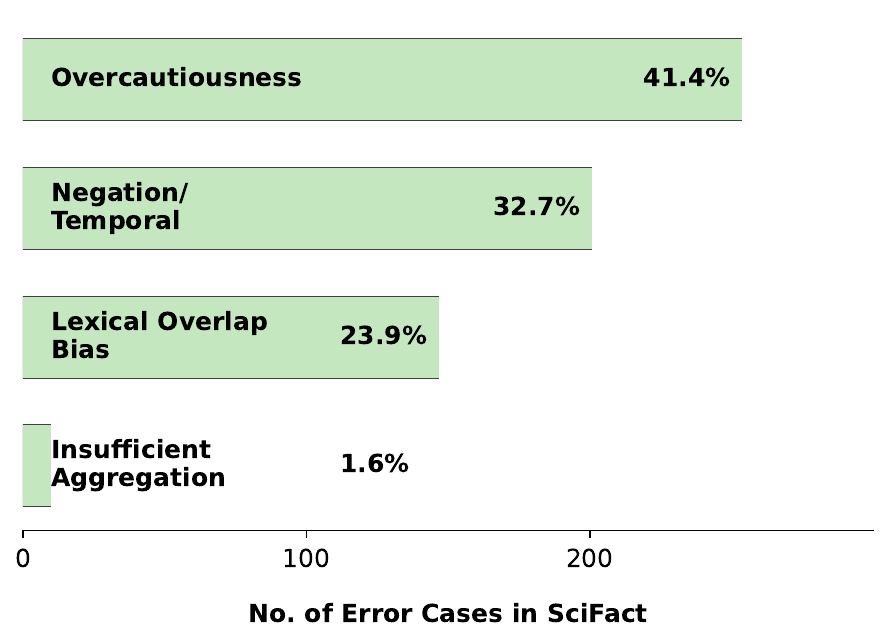}
        \caption{SciFact}
        \label{fig:error_scifact}
    \end{subfigure}
    \hfill
    \begin{subfigure}[b]{0.32\textwidth}
        \includegraphics[width=\textwidth]{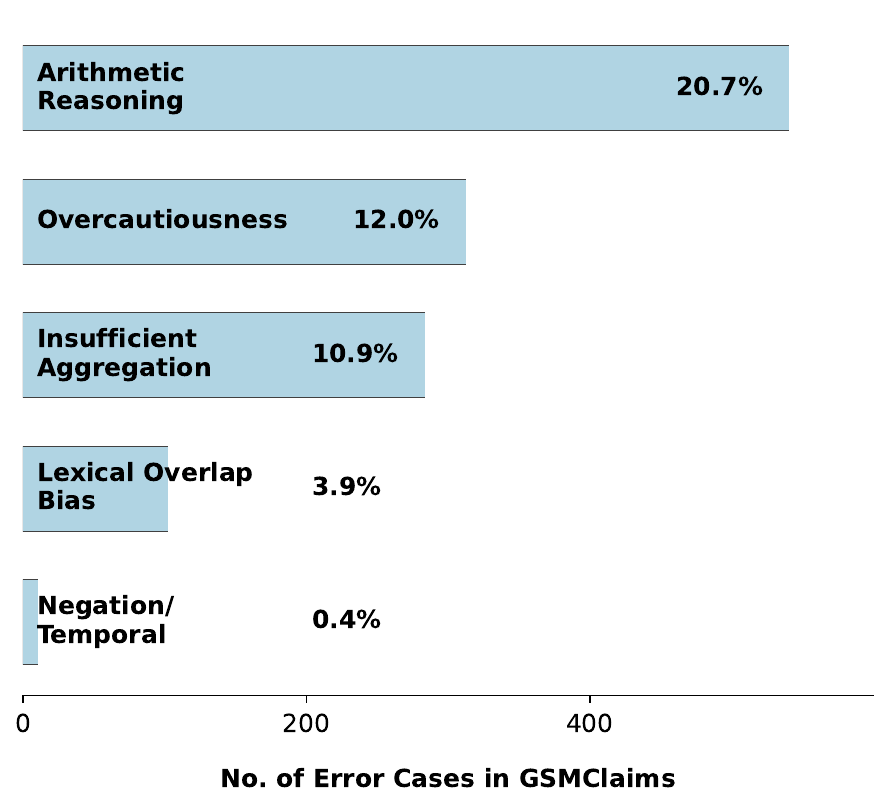}
        \caption{GSMClaims}
        \label{fig:error_gsm}
    \end{subfigure}
    \vspace{-0.7em}
    \caption{Distribution of error types on (a) LLMAggreFact, (b) SciFact, and (c) a math-reasoning benchmark derived from GSM8K. Error profiles vary dramatically across domains: general-domain verification is dominated by lexical overlap bias, scientific verification by overcautiousness, and mathematical verification by arithmetic reasoning failures.}
    \label{fig:combined_error_analysis}
    \vspace{-1em}
\end{figure*}

Error profiles vary dramatically by domain (Figure~\ref{fig:combined_error_analysis}). In LLMAggreFact, \textbf{Lexical Overlap Bias} is the most prevalent error (5.3\% of all examples), followed by \textbf{Insufficient Aggregation} (4.6\%) -- reflecting the dominance of direct evidence extraction in these datasets. In SciFact, \textbf{Overcautiousness} dominates (41.4\% of errors), with the verifier predicting ``not supported'' unless every component of a scientific claim is explicitly stated; \textbf{Negation/Temporal} errors are also substantial (32.8\%). In the mathematical domain, \textbf{Arithmetic Reasoning} errors are most common (43.2\% of errors), followed by \textbf{Overcautiousness} (25.0\%) and \textbf{Insufficient Aggregation} (22.7\%).

\subsection{Key Error Examples}

\paragraph{Lexical Overlap Bias (general domain):}
\begin{quote}
\small
\textbf{Claim:} Roberto Martinez felt Seamus Coleman should have been awarded a free-kick before the defender conceded the penalty that allowed Swansea to pinch a 1-1 draw at the Liberty Stadium. \\
\textbf{Ground Truth:} NO \quad \textbf{Predicted:} YES \\
\textbf{Analysis:} ClaimTrace matches surface phrases without verifying true entailment.
\end{quote}

\paragraph{Overcautiousness (scientific domain):}
\begin{quote}
\small
\textbf{Claim:} 1,000 genomes project enables mapping of genetic sequence variation consisting of rare variants with larger penetrance effects than common variants. \\
\textbf{Ground Truth:} YES \quad \textbf{Predicted:} NO \\
\textbf{Analysis:} The document discusses the identification of common variants and implications of rare variants, but ClaimTrace rejects the claim because the specific number of genomes is not mentioned.
\end{quote}

\paragraph{Arithmetic Reasoning (mathematical domain):}
\begin{quote}
\small
\textbf{Claim:} Janet makes \$18 every day at the farmers' market. \\
\textbf{Ground Truth:} YES \quad \textbf{Predicted:} NO \\
\textbf{Analysis:} The document provides the information needed to calculate the earnings, but ClaimTrace fails to execute the computation correctly.
\end{quote}

\section{Findings and Implications}

Our analysis yields three key findings with direct implications for benchmark design.

\paragraph{Finding 1: High scores primarily reflect retrieval and entailment.}
The dominance of direct evidence extraction (Pattern A) across LLMAggreFact means that a model achieving high balanced accuracy on this benchmark is primarily demonstrating its ability to locate and match relevant text spans. While this is a necessary component of verification, it is not sufficient for claims requiring deeper reasoning. The strong correlation between benchmark performance and Pattern A prevalence suggests that current metrics overestimate general verification ability.

\paragraph{Finding 2: Multi-sentence synthesis and numerical reasoning are under-tested.}
Patterns D (synthesis) and F (step-by-step verification) together account for a small fraction of the reasoning demands in LLMAggreFact. This means that verification systems can achieve competitive scores while having weak capabilities in exactly the reasoning types needed for complex, real-world claims. The near-random performance of both a 7B specialized verifier and our compact model on mathematical claims further demonstrates this gap.

\paragraph{Finding 3: Error profiles are domain-specific.}
The dramatic variation in error distributions across general, scientific, and mathematical domains shows that no single mitigation strategy suffices. Improving lexical overlap handling (the dominant general-domain error) would have minimal impact on scientific claim verification, where overcautiousness is the primary issue.

\paragraph{Recommendations.}
Based on these findings, we suggest the following directions for benchmark development: (1)~adversarial data mining that specifically targets lexical overlap to reduce surface-matching shortcuts; (2)~inclusion of more multi-hop claims requiring evidence synthesis across sentences; (3)~dedicated numerical reasoning components, since current benchmarks almost entirely lack such demands; (4)~domain-stratified evaluation that reports per-domain performance rather than single aggregate scores; and (5)~finer-grained labels such as \textsc{Partial} support to capture the nuanced nature of real-world claim verification.

\section{Conclusion and Limitations}

We presented a systematic analysis of the reasoning demands in claim verification benchmarks, generating and analyzing 24.1K structured reasoning traces across nine datasets. Our taxonomy reveals that direct evidence extraction dominates current benchmarks, while multi-sentence synthesis and numerical reasoning are severely under-represented. Domain-specific error analysis using a compact verifier shows that failure modes vary dramatically across general, scientific, and mathematical claims. These findings suggest that current high benchmark scores primarily reflect retrieval-plus-entailment ability rather than robust verification reasoning.

\paragraph{Limitations.}
Our analysis relies on reasoning traces from a single generator (GPT-4o-mini); different models may produce traces emphasizing different strategies. The analysis is limited to the nine datasets aggregated in LLMAggreFact. Manual coding of reasoning patterns, while conducted with stratified sampling and iterative consolidation, involves subjective judgment. The compact verifier used for error analysis has limited capacity, and a larger model might exhibit different error patterns.

\section*{Acknowledgments}
This research was developed with funding from the Defense Advanced Research Projects Agency's (DARPA) SciFy program (Agreement No. HR00112520300). The views expressed are those of the author and do not reflect the official policy or position of the Department of Defense or the U.S. Government.

\bibliographystyle{splncs04}
\bibliography{custom}

@article{thorne2018fever,
  title={FEVER: a large-scale dataset for fact extraction and VERification},
  author={Thorne, James and Vlachos, Andreas and Christodoulopoulos, Christos and Mittal, Arpit},
  journal={arXiv preprint arXiv:1803.05355},
  year={2018}
}

@inproceedings{yao2023react,
  title={React: Synergizing reasoning and acting in language models},
  author={Yao, Shunyu and Zhao, Jeffrey and Yu, Dian and Du, Nan and Shafran, Izhak and Narasimhan, Karthik and Cao, Yuan},
  booktitle={International Conference on Learning Representations (ICLR)},
  year={2023}
}

@article{jacovi2024chain,
  title={A chain-of-thought is as strong as its weakest link: A benchmark for verifiers of reasoning chains},
  author={Jacovi, Alon and Bitton, Yonatan and Bohnet, Bernd and Herzig, Jonathan and Honovich, Or and Tseng, Michael and Collins, Michael and Aharoni, Roee and Geva, Mor},
  journal={arXiv preprint arXiv:2402.00559},
  year={2024}
}

@article{wei2022chain,
  title={Chain-of-thought prompting elicits reasoning in large language models},
  author={Wei, Jason and Wang, Xuezhi and Schuurmans, Dale and Bosma, Maarten and Xia, Fei and Chi, Ed and Le, Quoc V and Zhou, Denny and others},
  journal={Advances in neural information processing systems},
  volume={35},
  pages={24824--24837},
  year={2022}
}

@article{kamoi2023wice,
  title={Wice: Real-world entailment for claims in wikipedia},
  author={Kamoi, Ryo and Goyal, Tanya and Rodriguez, Juan Diego and Durrett, Greg},
  journal={arXiv preprint arXiv:2303.01432},
  year={2023}
}

@article{wang2023factcheck,
  title={Factcheck-bench: Fine-grained evaluation benchmark for automatic fact-checkers},
  author={Wang, Yuxia and Reddy, Revanth Gangi and Mujahid, Zain Muhammad and Arora, Arnav and Rubashevskii, Aleksandr and Geng, Jiahui and Afzal, Osama Mohammed and Pan, Liangming and Borenstein, Nadav and Pillai, Aditya and others},
  journal={arXiv preprint arXiv:2311.09000},
  year={2023}
}

@article{malaviya2023expertqa,
  title={Expertqa: Expert-curated questions and attributed answers},
  author={Malaviya, Chaitanya and Lee, Subin and Chen, Sihao and Sieber, Elizabeth and Yatskar, Mark and Roth, Dan},
  journal={arXiv preprint arXiv:2309.07852},
  year={2023}
}

@article{liu2023evaluating,
  title={Evaluating verifiability in generative search engines},
  author={Liu, Nelson F and Zhang, Tianyi and Liang, Percy},
  journal={arXiv preprint arXiv:2304.09848},
  year={2023}
}

@article{zhu2021mediasum,
  title={MediaSum: A large-scale media interview dataset for dialogue summarization},
  author={Zhu, Chenguang and Liu, Yang and Mei, Jie and Zeng, Michael},
  journal={arXiv preprint arXiv:2103.06410},
  year={2021}
}

@article{hu2023meetingbank,
  title={MeetingBank: A benchmark dataset for meeting summarization},
  author={Hu, Yebowen and Ganter, Tim and Deilamsalehy, Hanieh and Dernoncourt, Franck and Foroosh, Hassan and Liu, Fei},
  journal={arXiv preprint arXiv:2305.17529},
  year={2023}
}

@article{narayan2018don,
  title={Don't give me the details, just the summary! topic-aware convolutional neural networks for extreme summarization},
  author={Narayan, Shashi and Cohen, Shay B and Lapata, Mirella},
  journal={arXiv preprint arXiv:1808.08745},
  year={2018}
}

@article{nallapati2016abstractive,
  title={Abstractive text summarization using sequence-to-sequence rnns and beyond},
  author={Nallapati, Ramesh and Zhou, Bowen and Gulcehre, Caglar and Xiang, Bing and others},
  journal={arXiv preprint arXiv:1602.06023},
  year={2016}
}

@article{tang2022understanding,
  title={Understanding factual errors in summarization: Errors, summarizers, datasets, error detectors},
  author={Tang, Liyan and Goyal, Tanya and Fabbri, Alexander R and Laban, Philippe and Xu, Jiacheng and Yavuz, Semih and Kry{\'s}ci{\'n}ski, Wojciech and Rousseau, Justin F and Durrett, Greg},
  journal={arXiv preprint arXiv:2205.12854},
  year={2022}
}

@misc{hu2021lora,
      title={LoRA: Low-Rank Adaptation of Large Language Models}, 
      author={Edward J. Hu and Yelong Shen and Phillip Wallis and Zeyuan Allen-Zhu and Yuanzhi Li and Shean Wang and Lu Wang and Weizhu Chen},
      year={2021},
      eprint={2106.09685},
      archivePrefix={arXiv},
      primaryClass={cs.CL},
      url={https://arxiv.org/abs/2106.09685}, 
}

@misc{deepseek-r1-2025,
      title={DeepSeek-R1: Incentivizing Reasoning Capability in LLMs via Reinforcement Learning}, 
      author={DeepSeek-AI and Daya Guo and Dejian Yang and Haowei Zhang and Junxiao Song and Ruoyu Zhang and Runxin Xu and Qihao Zhu and Shirong Ma and Peiyi Wang and Xiao Bi and Xiaokang Zhang and Xingkai Yu and Yu Wu and Z. F. Wu and Zhibin Gou and Zhihong Shao and Zhuoshu Li and Ziyi Gao and Aixin Liu and Bing Xue and Bingxuan Wang and Bochao Wu and Bei Feng and Chengda Lu and Chenggang Zhao and Chengqi Deng and Chenyu Zhang and many more},
      year={2025},
      eprint={2501.12948},
      archivePrefix={arXiv},
      primaryClass={cs.CL},
      url={https://arxiv.org/abs/2501.12948}, 
}

@misc{cobbe2021,
      title={Training Verifiers to Solve Math Word Problems}, 
      author={Karl Cobbe and Vineet Kosaraju and Mohammad Bavarian and Mark Chen and Heewoo Jun and Lukasz Kaiser and Matthias Plappert and Jerry Tworek and Jacob Hilton and Reiichiro Nakano and Christopher Hesse and John Schulman},
      year={2021},
      eprint={2110.14168},
      archivePrefix={arXiv},
      primaryClass={cs.LG},
      url={https://arxiv.org/abs/2110.14168}, 
}

@misc{zha2023alignscore,
      title={AlignScore: Evaluating Factual Consistency with a Unified Alignment Function}, 
      author={Yuheng Zha and Yichi Yang and Ruichen Li and Zhiting Hu},
      year={2023},
      eprint={2305.16739},
      archivePrefix={arXiv},
      primaryClass={cs.CL},
      url={https://arxiv.org/abs/2305.16739}, 
}

@inproceedings{wadden-etal-2020-fact,
    title = "Fact or Fiction: Verifying Scientific Claims",
    author = "Wadden, David  and
      Lin, Shanchuan  and
      Lo, Kyle  and
      Wang, Lucy Lu  and
      van Zuylen, Madeleine  and
      Cohan, Arman  and
      Hajishirzi, Hannaneh",
    editor = "Webber, Bonnie  and
      Cohn, Trevor  and
      He, Yulan  and
      Liu, Yang",
    booktitle = "Proceedings of the 2020 Conference on Empirical Methods in Natural Language Processing (EMNLP)",
    month = nov,
    year = "2020",
    address = "Online",
    publisher = "Association for Computational Linguistics",
    url = "https://aclanthology.org/2020.emnlp-main.609/",
    doi = "10.18653/v1/2020.emnlp-main.609",
    pages = "7534--7550",
}

@misc{gemmateam2025gemma3technicalreport,
      title={Gemma 3 Technical Report}, 
      author={GemmaTeam and Aishwarya Kamath and Johan Ferret and Shreya Pathak and Nino Vieillard and Ramona Merhej and Sarah Perrin and Tatiana Matejovicova and Alexandre Ramé and Morgane Rivière and Louis Rouillard and Thomas Mesnard and Geoffrey Cideron and Jean-bastien Grill and Sabela Ramos and Edouard Yvinec and Michelle Casbon and Etienne Pot and Ivo Penchev and Gaël Liu and Francesco Visin and Kathleen Kenealy and Lucas Beyer and Xiaohai Zhai and Anton Tsitsulin and Robert Busa-Fekete and Alex Feng and Noveen Sachdeva and Benjamin Coleman and many more},
      year={2025},
      eprint={2503.19786},
      archivePrefix={arXiv},
      primaryClass={cs.CL},
      url={https://arxiv.org/abs/2503.19786}, 
}

@misc{tang2024minicheck,
      title={MiniCheck: Efficient Fact-Checking of LLMs on Grounding Documents}, 
      author={Liyan Tang and Philippe Laban and Greg Durrett},
      year={2024},
      eprint={2404.10774},
      archivePrefix={arXiv},
      primaryClass={cs.CL},
      url={https://arxiv.org/abs/2404.10774}, 
}

@InProceedings{pmlr-v139-zhao21c,
  title = 	 {Calibrate Before Use: Improving Few-shot Performance of Language Models},
  author =       {Zhao, Zihao and Wallace, Eric and Feng, Shi and Klein, Dan and Singh, Sameer},
  booktitle = 	 {Proceedings of the 38th International Conference on Machine Learning},
  pages = 	 {12697--12706},
  year = 	 {2021},
  editor = 	 {Meila, Marina and Zhang, Tong},
  volume = 	 {139},
  series = 	 {Proceedings of Machine Learning Research},
  month = 	 {18--24 Jul},
  publisher =    {PMLR},
  url = 	 {https://proceedings.mlr.press/v139/zhao21c.html}
}

\end{CJK*}
\end{document}